\documentclass[sigconf]{acmart}
\AtBeginDocument{%
  }

\copyrightyear{2026}
\acmYear{2026}
\setcopyright{cc}
\setcctype{by}
\acmDOI{10.1145/3767308.3836287}
\acmConference[MM '26]
  {Proceedings of the 34th ACM International Conference on Multimedia}
  {November 10--14, 2026}
  {Rio de Janeiro, Brazil}
\acmBooktitle{Proceedings of the 34th ACM International Conference on Multimedia
  (MM '26), November 10--14, 2026, Rio de Janeiro, Brazil}
\acmISBN{979-8-4007-2213-4/2026/11}
\usepackage{tabularx}
\usepackage{multirow}
\begin{document}

\title{SynerGest: Temporal-Spatial Synergistic Gesture Generation for Dyadic Conversations}

\author{Chengxiang Guo}
\additionalaffiliation{%
  \institution{Yunnan Key Laboratory of Artificial Intelligence}
  \city{Kunming}
  \country{China}}
\affiliation{%
  \department{Faculty of Information Engineering and Automation}
  \institution{Kunming University of Science and Technology}
  \city{Kunming}
  \country{China}}
\email{20242204355@stu.kust.edu.cn}

\author{Jian Wang}
\correspondingauthor
\authornotemark[1]
\affiliation{%
  \department{Faculty of Information Engineering and Automation}
  \institution{Kunming University of Science and Technology}
  \city{Kunming}
  \country{China}}
\email{jianwang@kust.edu.cn}

\author{Lanmiao Liu}
\affiliation{%
  \institution{Utrecht University}
  \city{Utrecht}
  \country{Netherlands}}
\affiliation{%
  \institution{The Max Planck Institute for Psycholinguistics}
  \city{Nijmegen}
  \country{Netherlands}}
\email{lanmiao.liu@mpi.nl}

\author{Zhan Tang}
\affiliation{%
  \institution{Zhejiang University}
  \city{Hangzhou}
  \country{China}}
\email{erictz@zju.edu.cn}

\author{Junhua Fei}
\authornotemark[1]
\affiliation{%
  \department{Faculty of Information Engineering and Automation}
  \institution{Kunming University of Science and Technology}
  \city{Kunming}
  \country{China}}
\email{20242204125@stu.kust.edu.cn}

\author{Xiao Li}
\authornotemark[1]
\affiliation{%
  \department{Faculty of Information Engineering and Automation}
  \institution{Kunming University of Science and Technology}
  \city{Kunming}
  \country{China}}
\email{20252204123@stu.kust.edu.cn}

\renewcommand{\shortauthors}{Chengxiang Guo et al.}

\begin{abstract}
Co-speech gesture generation in dyadic conversations for creating immersive virtual avatars remains highly challenging. Existing models often struggle to balance kinematic stability with rhythmic responsiveness and suffer from semantic entanglement, leading to either generic mean-pose regression or unnatural micro-jitters. To address these challenges, we propose SynerGest, a high-fidelity dyadic gesture generation framework grounded in Contrastive Flow Matching (CFM). Specifically, our CFM objective explicitly penalizes mismatched semantic pairs during training, discouraging spatial shortcuts and generic, over-smoothed predictions while improving conditional semantic alignment. Furthermore, to overcome architectural bottlenecks, we introduce the Impulse-Trend Synergistic Attention (ITSA) module. Grounded in the key insight of decoupling "what moves" from "when it moves," ITSA models sharp, speech-driven rhythmic impulses separately from continuous kinematic trends, supporting precise audio-motion synchronization and global structural stability. Extensive qualitative and quantitative evaluations on two public datasets demonstrate that SynerGest achieves state-of-the-art performance, delivering contextually responsive and kinematically fluid dyadic interactions.
\end{abstract}


\begin{teaserfigure}
  \centering
  \includegraphics[width=\textwidth, height=6.5cm, keepaspectratio]{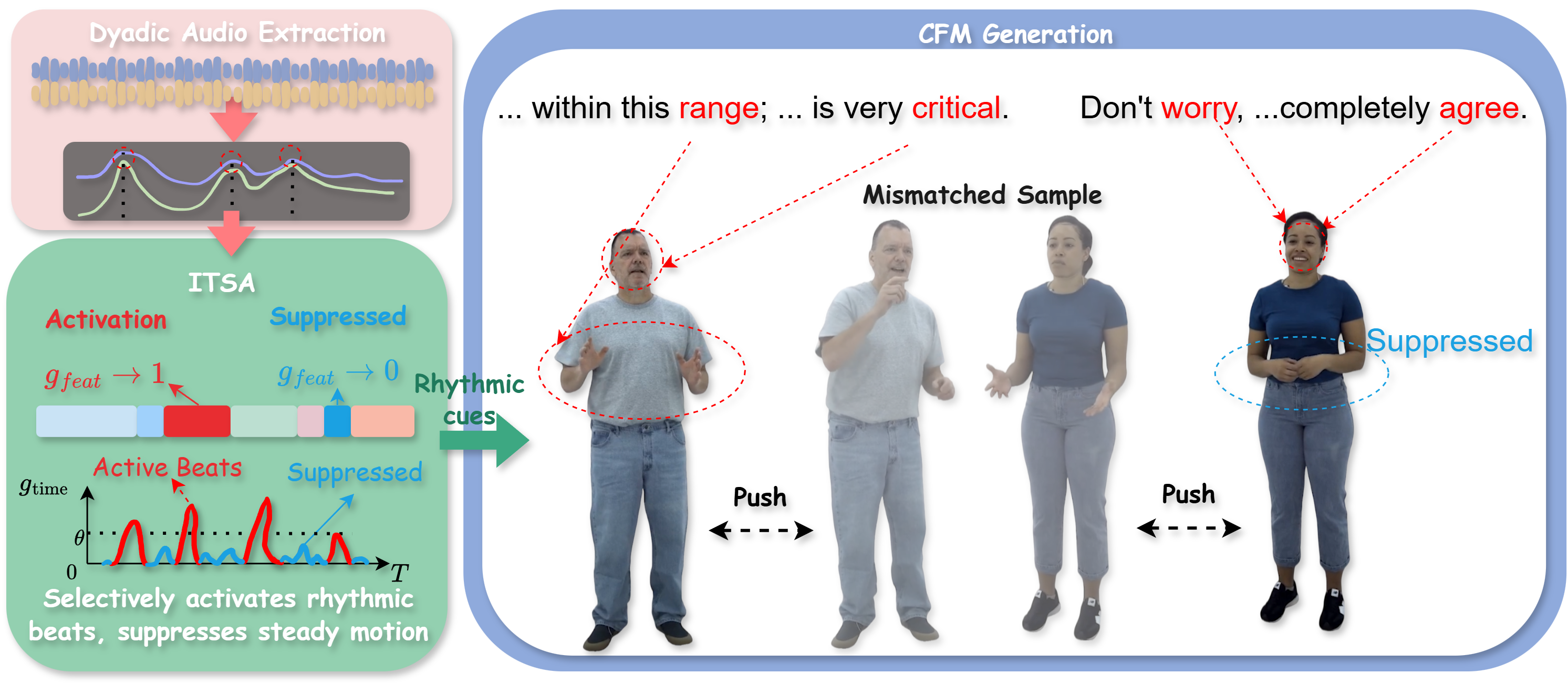}
  \caption{Overview of \textbf{SynerGest}. ITSA selects rhythmic motion cues, while CFM separates mismatched audio--motion pairs to preserve synchrony and motion diversity.}
  \Description{A conceptual diagram of the SynerGest framework divided into left and right panels. The left panel shows 'Dyadic Audio Extraction' and 'ITSA' for spatial and temporal gating. The right panel shows 'CFM Generation' where a ghosted 'Mismatched Sample' is pushed away to generate highly synchronized active gestures and suppress resting trends.}
  \label{fig:teaser}
\end{teaserfigure}


\maketitle

\section{Introduction}
\label{sec:intro}

Nonverbal communication encompassing facial expressions, body language, and hand gestures is integral to natural human interaction, particularly in dyadic (two-person) conversations, where it enriches the exchange with crucial contextual cues and mutual emotional depth~\cite{McNeill1992Hand}. Endowing digital avatars with the holistic capabilities required for natural dyadic interaction has become a cornerstone of state-of-the-art research~\cite{Ng2024Audio2Photoreal,Agrawal2024Seamless}. While audio-driven animation has advanced, mapping acoustic signals directly to parametric body models frequently produces kinematically constrained and uncanny results, lacking the non-verbal subtleties essential for realistic interaction. Co-speech motion combines continuously occurring, rhythm-related movement driven by speech cadence~\cite{Alexanderson2020Style} with sparse, semantics-aware gestures~\cite{Zhang2024Semantic,Zhang2025SemTalk,liu2025semges} that are infrequent but critical for conveying specific intent. In dyadic conversation, ensuring the precise semantic alignment of these sparse gestures is even more demanding than in monologues, as they must continuously resonate with the conversational context. The listener is not merely a passive recipient but an active participant who shapes the narrative flow through continuous non-verbal feedback. Learning an effective correspondence from latent acoustic features to this high-dimensional, spatiotemporally coherent interaction manifold remains a fundamental bottleneck in current generative frameworks.

Current research remains constrained by significant limitations regarding temporal consistency and spatial resolution. Broadly, the vast majority of holistic motion generation frameworks~\cite{Liu2024EMAGE, Chen2024DiffSHEG, Yi2023Holistic,liu2026holisticsemges} are monadic. They generate one person's motion without explicitly conditioning on the behavior or speech of a conversational partner, thereby omitting the joint dynamics of human communication. Authentic dyadic conversation constitutes a dynamic feedback loop, where communication is a jointly-constructed process. The listener's reactive behaviors, ranging from subtle postural shifts to affective facial cues, are integral components of the spatiotemporal interaction manifold. Authentic communication requires the digital avatar to not only project its own intent but also continuously adapt to the evolving behavioral cues of its partner. Monadic approaches fundamentally struggle to capture this cross-modal responsiveness.

To bridge this gap, research has shifted toward dyadic embodiment, catalyzed by pioneering frameworks and large-scale datasets such as \textit{Audio2Photoreal}~\cite{Ng2024Audio2Photoreal} and \textit{Seamless}~\cite{Agrawal2024Seamless}. At a technical level, Audio2Photoreal combines autoregressive guide-pose prediction with diffusion-based detail synthesis, whereas Seamless employs a Diffusion Transformer trained with a flow-matching objective and multi-step ODE inference. These multi-stage or iterative pipelines impose substantial computational overhead. Recent monadic work such as GestureLSM~\cite{Liu2025GestureLSM} adopts Flow Matching (FM)~\cite{Lipman2023FlowMatching} for efficiency but relies on a standard conditional FM objective. This objective learns from matched condition-trajectory pairs without explicitly contrasting condition mismatches, which can limit sensitivity to ambiguous conversational audio. Additionally, across both diffusion and FM paradigms, sequence modeling heavily relies on standard Spatial-Temporal Transformers~\cite{Vaswani2017Attention,Aksan2021Spatio}. This introduces a computational bottleneck because full self-attention has quadratic complexity in sequence length~\cite{Vaswani2017Attention}. Conventional attention also does not explicitly distinguish the frequency characteristics of high-frequency, speech-driven transients and low-frequency kinematic trajectories, motivating a frequency-aware token mixer.

To address these multifaceted challenges, we propose \textbf{SynerGest}, a novel framework for dyadic photoreal embodiment (\textbf{Fig.~\ref{fig:teaser}}). Specifically, to tackle the tendency of vanilla FM to produce semantically under-constrained trajectories under ambiguous audio, we adapt \textbf{Contrastive Flow Matching (CFM)} to dual-audio dyadic conditioning. Building upon foundational FM~\cite{Lipman2023FlowMatching}, Contrastive Flow Matching~\cite{Stoica2025ContrastiveFM}, and its recent use in monadic co-speech gesture generation~\cite{liu2026holisticsemges}, our objective compares matched and shuffled audio conditions at the same latent state and maximizes the distance between their predicted velocity fields. This formulation encourages condition-sensitive trajectories and reduces the tendency to produce generic, over-smoothed gestures under ambiguous conversational contexts while preserving the efficiency of ODE-based generation.

Second, to overcome the semantic entanglement of standard Transformers, we propose the \textbf{Impulse-Trend Synergistic Attention (ITSA)} mechanism. By replacing quadratic self-attention with pooling-based gating streams, the front-end of ITSA explicitly decouples motion into an \textit{Impulse Temporal Gating} stream (``when it moves'') and a \textit{Trend Feature Gating} stream (``what moves''). Crucially, rendering kinematically plausible motion requires more than local spatiotemporal decoupling; the generated motion must inherently align with the complex spatiotemporal distribution of natural human motion. To achieve this, the ITSA architecture intrinsically incorporates a standard Multi-Head Attention (MHA) layer as its final global refinement stage. While the dual-gating streams provide efficient local feature disentanglement, this embedded MHA captures the global contextual dependencies required for long-term trajectory planning. This unified hybrid design reduces the tendency of the downstream generator to produce out-of-distribution jitter, supporting the kinematic realism and global stability of the synthesized conversational gestures.

Our contributions are as follows:
\begin{itemize}
    \item \textbf{From Isolated to Interaction-Aware Synthesis.} We introduce a dyadic-conditioned framework for conversational motion generation. Rather than focusing on isolated, unilateral generation driven by a single audio source, our model explicitly leverages dual-audio context to synthesize highly responsive behaviors. 
    \item \textbf{Contrastive Flow Matching for Robust Generation.} Moving beyond standard DMs and vanilla FM, our dyadic contrastive flow objective explicitly penalizes mismatched audio conditions, improving conditional separation and continuous audio-motion alignment.
    \item \textbf{ITSA with Global Refinement.} We introduce a novel, highly efficient token mixer that replaces standard self-attention with pooling-based dual-gating streams. ITSA explicitly disentangles sharp rhythmic impulses from smooth kinematic trends, while its integrated MHA backend ensures long-term global stability.
    \item \textbf{State-of-the-Art Performance.} Extensive experiments on the Audio2Photoreal and Seamless Interaction datasets demonstrate that SynerGest significantly outperforms existing baselines in temporal consistency and inference efficiency.
\end{itemize}

\section{Related Work}
\label{sec:related_work}

\subsection{From Monadic to Dyadic Co-Speech Generation}
The synthesis of animated characters has progressed from speech-driven 2D lip synchronization~\cite{Prajwal2020Wav2Lip} and portrait-avatar animation~\cite{Cao2022StyleAvatar} to expressive 3D facial and full-body motion. Early data-driven body-language systems relied on prosody-conditioned Hidden Markov Models~\cite{Levine2009}, which limited the expressiveness and naturalness of synthesized motion~\cite{Nyatsanga2023Comprehensive}. Subsequent facial-animation methods use autoregressive Transformers~\cite{Fan2022Faceformer}, neural radiance fields~\cite{Guo2021ADNeRF}, or explicit 3D motion coefficients~\cite{Zhang2023SadTalker}. For body motion, Text2Gestures generates emotive gestures from natural-language descriptions~\cite{Bhattacharya2021Speech2Gesture}, while speech-driven normalizing flows model diverse gesticulation from acoustic input~\cite{Alexanderson2020Style}. Supported by large-scale multimodal motion-capture datasets such as BEAT~\cite{Liu2022BEAT}, a growing body of work targets co-speech gesture generation from multimodal contexts~\cite{Ginosar2019Gesture,Kucherenko2020Gesticulator,Qian2021SpeechTemplates,Yoon2020Speech,Yi2023Holistic,Liu2024EMAGE,Chen2024DiffSHEG}. GestureLSM~\cite{Liu2025GestureLSM} further applies Flow Matching in a quantized latent space for real-time generation. Despite this progress, these methods remain monadic and do not explicitly capture the mutual observations that define dyadic interaction.

Dyadic embodiment has been significantly advanced by integrated benchmarks such as \textit{Au\-dio\-2\-Pho\-to\-re\-al}~\cite{Ng2024Audio2Photoreal}, which provides both a dataset and a framework, and \textit{Seamless}~\cite{Agrawal2024Seamless}. Related work also models non-deterministic listener facial motion and non-verbal feedback~\cite{Ng2022Listening}. Audio2Photoreal combines autoregressive guide-pose prediction with diffusion-based detail synthesis, whereas Seamless uses a flow-matching-trained Diffusion Transformer with multi-step ODE inference. We reduce the resulting iterative cost with an eight-step flow-matching backbone tailored to dyadic generation.

\subsection{Generative Models and Spatial-Temporal Modeling}
Recent continuous generative modeling includes Denoising Diffusion Probabilistic Models (DDPMs)~\cite{Ho2020DDPM} and score-based stochastic differential equation models~\cite{Song2020Score}. Motion models such as MDM and MotionDiffuse demonstrate diverse text-conditioned motion generation~\cite{Tevet2022MDM,Zhang2022MotionDiffuse}. Their sampling is iterative, however, and MDM reports roughly 1,000 network evaluations for a generated sample. Consequently, Flow Matching (FM) and related rectified-flow formulations~\cite{Lipman2023FlowMatching,Liu2023RectifiedFlow} have emerged as efficient alternatives. They train continuous velocity fields without simulating the learned ODE during training and can map a simple prior to the data distribution through an ODE~\cite{Albergo2023Building}. Nevertheless, standard conditional FM objectives learn from matched condition-trajectory pairs and lack an explicit mechanism for contrasting condition mismatches. This can weaken conditional separation when conversational audio is ambiguous.

Furthermore, across both diffusion and FM paradigms, sequence modeling typically relies on standard Transformers~\cite{Fan2022Faceformer,Zhang2023T2MGPT,Yi2023Holistic}. For dense 30 fps motion generation, modeling extended conversational contexts creates severe architectural bottlenecks because full self-attention grows quadratically with sequence length~\cite{Vaswani2017Attention}. While efficient sequence models such as linear attention~\cite{Choromanski2020Performer} and state-space models~\cite{Gu2023Mamba} mitigate computational cost, they do not explicitly distinguish smooth kinematic trends from sharp rhythmic transients. To address this representational limitation, we augment global Multi-Head Attention (MHA) with lightweight pooling-based dual-gating streams that model these frequency characteristics separately. The gating streams add little overhead, while MHA is retained for global refinement, as detailed in Sec.~\ref{subsec:itsa}.

\subsection{Cross-Modal Alignment and Contrastive Learning}
Precise temporal alignment between speech and facial motion is central to talking-face synthesis~\cite{Peng2024SyncTalk,Chung2016SyncNet}, and analogous alignment is required between speech and body motion in gesture generation. Because speech-to-gesture mapping is inherently one-to-many, regression-based objectives can favor averaged motions and yield over-smoothed results~\cite{Nyatsanga2023Comprehensive}. Recent works emphasize separating periodic rhythm from semantic intent to combat this over-smoothing~\cite{Ao2022Rhythmic,Liu2022DisCo,Zhang2024Semantic,Zhang2025SemTalk}. Nevertheless, these methods primarily introduce representation or architectural mechanisms, whereas we modify the conditional flow objective itself to distinguish matched and mismatched conversational conditions.

In representation learning, Contrastive Learning~\cite{Chen2020SimCLR,Radford2021CLIP,Oord2018InfoNCE} is recognized for building robust, discriminative feature spaces. Its efficacy extends to audio-language~\cite{Elizalde2023CLAP} and audio-visual representation learning~\cite{Afouras2022Self,Rouditchenko2020AVLnet}. Contrastive Flow Matching separates conditional flows by introducing negative trajectories~\cite{Stoica2025ContrastiveFM}, and SemConFlow applies this principle to monadic holistic gesture generation using audio-text conditions and mismatched target motions~\cite{liu2026holisticsemges}. Our dyadic variant instead conditions on the two conversational audio streams and compares predicted velocities under matched and shuffled audio conditions at the same latent state. The resulting Audio-Motion Contrastive Loss encourages the ODE trajectory to remain sensitive to the correct conversational context while preserving kinematic stability.

\section{Method}
\label{sec:method}



\begin{figure*}[htbp]
  \centering
  \includegraphics[width=\textwidth, height=7.5cm, keepaspectratio]{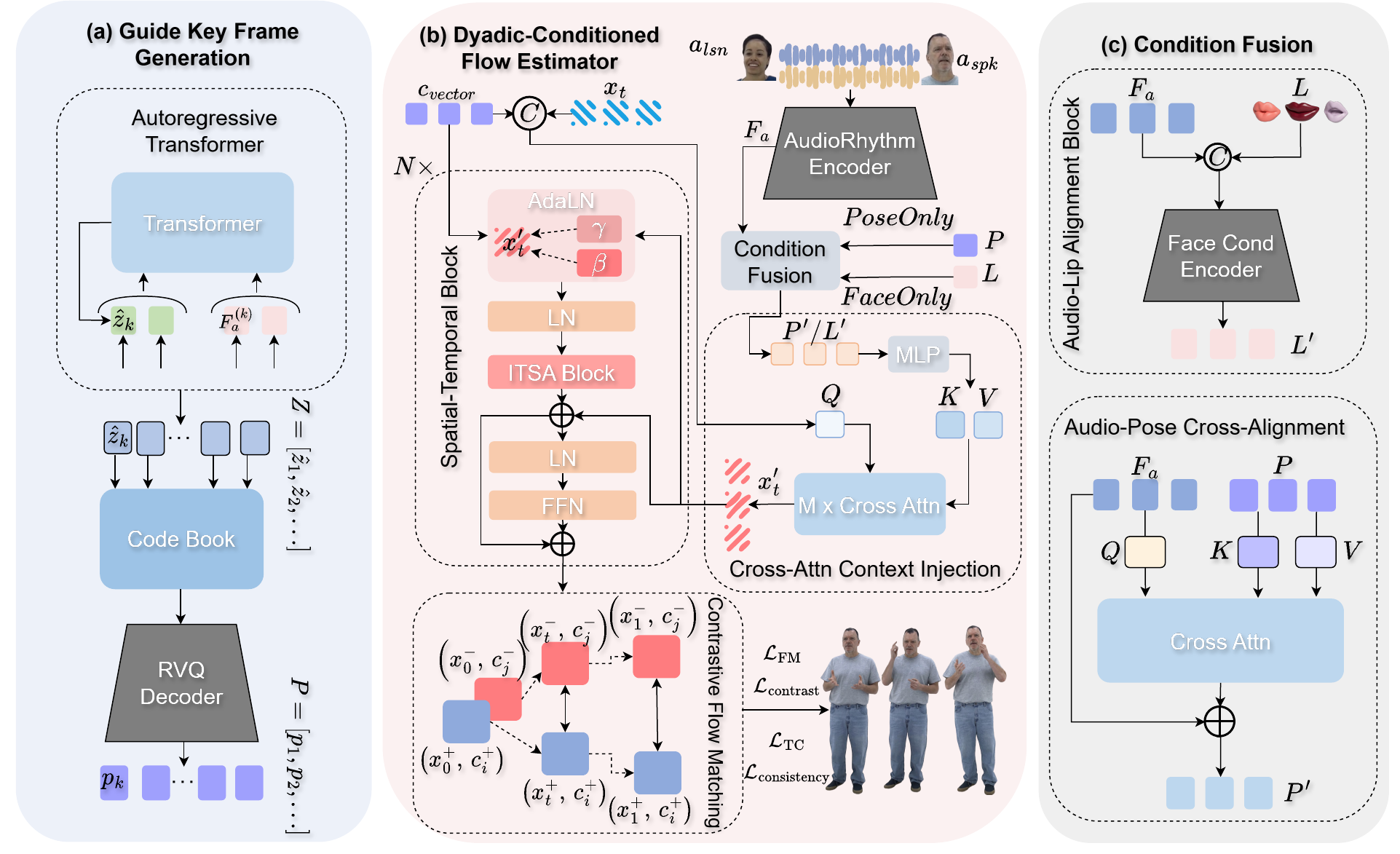} 
  \caption{Architecture of \textbf{SynerGest}: \textbf{(a)} RVQ-based guide key-frame generation, \textbf{(b)} the ITSA-equipped dyadic flow estimator, and \textbf{(c)} audio--lip and audio--pose condition fusion.}
  \label{fig:architecture} 
  \Description{A detailed three-panel architecture diagram of the SynerGest framework featuring dyadic-conditioned flow matching.}
\end{figure*}

\subsection{Overview}

We propose \textbf{SynerGest}, a high-fidelity co-speech gesture generator built on Contrastive Flow Matching (CFM), designed to model the interaction within a dyadic conversation. The generator synthesizes motion $\mathbf{x} \in \mathbb{R}^{T \times D}$ for the target avatar (acting as the listener in our joint-modeling scenario), where $T$ denotes the sequence length and $D$ represents the modality-specific spatial dimension. We first extract raw audio waveforms from both the target listener ($\mathbf{a}_{lsn}$) and the conversational partner ($\mathbf{a}_{spk}$). These are denoted as $\mathbf{A} = (\mathbf{a}_{lsn}, \mathbf{a}_{spk})$ and jointly condition generation. A shared, pretrained \textit{wav2vec 2.0} backbone combined with mel-spectrogram features processes the two channels independently. To accommodate distinct kinetic properties, the flow estimator separates body and facial-expression pathways (``PoseOnly'' and ``FaceOnly'' in \textbf{Fig.~\ref{fig:architecture}}). Rather than using a stochastic diffusion chain, we learn a step-size-conditioned velocity field $v_\theta(\mathbf{x}_t,t,d,\mathbf{c})$, where $t$ is the absolute flow time and $d$ is the requested integration step size. The field transports samples from a Gaussian prior $p_0(\mathbf{x})=\mathcal{N}(0,I)$ to the empirical distribution $p_1(\mathbf{x})$. For fixed initial noise and audio, integrating from $t=0$ to $1$ gives a deterministic ODE trajectory; drawing a fresh initial sample yields a different plausible gesture for the same audio.

\subsection{Conditional Feature Generation and Fusion}
\label{subsec:conditional_fusion}

As illustrated in our overall architecture, the conditional pipeline follows a structured progression from raw unimodal extraction to multimodal interactive alignment. First, the raw audio waveforms of the target listener ($\mathbf{a}_{lsn}$) and the speaker ($\mathbf{a}_{spk}$) are independently routed through the frozen \textit{AudioRhythm Encoder}. These embeddings are concatenated to jointly encode the dyadic conversational context, forming the base acoustic feature $\mathbf{F}_a$. Crucially, to accommodate their distinct kinetic properties, our framework adopts a decoupled extraction strategy for body pose and facial expressions. For body dynamics, a Guide Transformer~\cite{Ng2024Audio2Photoreal} autoregressively predicts discrete RVQ tokens from the audio stream, which are decoded to provide 1~fps initial pose keyframes $\mathbf{P}$ as a structural physical prior. Conversely, initial lip vertex features $\mathbf{L}$ are extracted via a pretrained lip-vertex regressor. These two modality-specific priors feed into independent generation pipelines: the \textit{PoseOnly} generator for body dynamics and the \textit{FaceOnly} generator for facial expressions.

These independently extracted preliminary features ($\mathbf{P}$ and $\mathbf{L}$) are subsequently aligned with the dyadic acoustic feature $\mathbf{F}_a$ within the \textit{Condition Fusion} block (\textbf{Fig.~\ref{fig:architecture}c}). Specifically, for body dynamics, an \textit{Audio-Pose Cross-Alignment} mechanism treats the acoustic feature $\mathbf{F}_a$ as the Query and the guide poses $\mathbf{P}$ as the Key and Value. Through this cross-attention operation, we extract the aligned pose semantic feature $\mathbf{P}'$. Concurrently, for facial expressions, an \textit{Audio-Lip Alignment} pathway concatenates the acoustic feature $\mathbf{F}_a$ with the lip vertex features $\mathbf{L}$ to produce the enhanced lip representation $\mathbf{L}'$. Ultimately, for each independent pipeline, these fused representations ($\mathbf{P}'$ or $\mathbf{L}'$) are processed through a dedicated lightweight MLP. The outputs of the MLP serve as the Key and Value for the subsequent cross-attention layers within their respective flow estimator backbones, ensuring dense semantic synchronization between the modality-specific priors and the generated trajectories.

\subsection{Dyadic-Conditioned Flow Estimator and Context Injection}
\label{subsec:flow_estimator}

At the core of our proposed architecture lies the \textbf{Dy\-ad\-ic-Con\-di\-tioned Flow Es\-ti\-ma\-tor}, which parameterizes the velocity field $v_\theta$ during training and functions as the derivative for the ordinary differential equation (ODE) solver during inference. As shown in the primary backbone of \textbf{Fig.~\ref{fig:architecture}b}, our network effectively integrates the conditional features into the state trajectory $\mathbf{x}_t$ through a structured context injection mechanism.

\paragraph{Condition Vector Formulation and Projection.} To parameterize the conditioning manifold for the flow estimator, the modality-specific condition tensors (derived from $\mathbf{P}'$ or $\mathbf{L}'$ via the MLP described in Sec.~\ref{subsec:conditional_fusion}) serve directly as the Keys ($\mathbf{K}$) and Values ($\mathbf{V}$) within the subsequent cross-attention layers. Concurrently, the unified conditioning vector $\mathbf{c}_{\text{vector}}$ encodes the absolute flow time $t$, the requested step size $d$, and the person-specific identity prior. This vector acts as the explicit control signal driving the Adaptive Layer Normalization (AdaLN) mechanism across all Spatial-Temporal Blocks.

\paragraph{Spatiotemporal Trajectory Modeling.} Within the primary backbone, the state trajectory $\mathbf{x}_t$ and the conditioning vector $\mathbf{c}_{\text{vector}}$ are concatenated (denoted as \textcircled{c}) and projected via a linear layer to form the initial Query ($\mathbf{Q}$). This query is routed through a cascade of $M=3$ stacked Cross-Attention blocks, where it actively attends to the projected multimodal conditions ($\mathbf{K}, \mathbf{V}$), achieving dense cross-modal semantic synchronization at the earliest integration stage. Subsequently, the conditionally enriched trajectory flows into $N=8$ stacked Spatial-Temporal Blocks. As presented in the dashed box of \textbf{Fig.~\ref{fig:architecture}b}, each block follows a standard conditional Transformer layout but replaces the vanilla self-attention with our proposed ITSA module. Specifically, the trajectory is first modulated by Adaptive Layer Normalization (AdaLN)~\cite{Peebles2023DiT}. AdaLN projects $\mathbf{c}_{\text{vector}}$ into affine parameters $(\gamma, \beta)$ to modulate the trajectory via $\mathbf{x}'_t = \gamma \cdot \text{LN}(\mathbf{x}_t) + \beta$, providing dynamic, condition-aware feature scaling. Following this, standard Layer Normalization (LN) is applied before both the ITSA module and the Feed-Forward Network (FFN), ensuring numerical stability during the intra-modal kinematic modeling and residual feature refinement. 

\begin{figure}[htbp]
  \centering
  \includegraphics[width=1\linewidth]{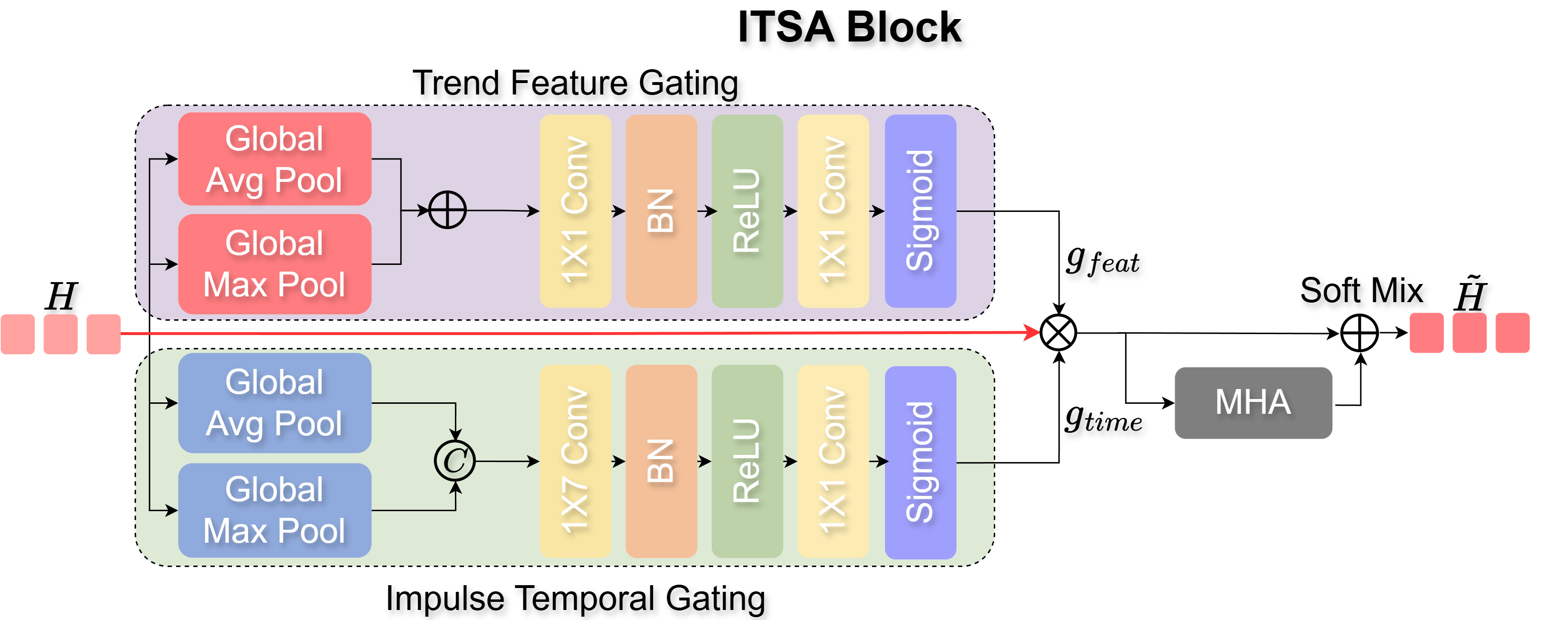} 
  \caption{ITSA decouples trend-feature and impulse-temporal gating, then adaptively fuses both branches with global multi-head attention.}
  \label{fig:itsa} 
  \Description{A block diagram showing the architecture of the Impulse-Trend Synergistic Attention module. It details the parallel Trend Feature Gating and Impulse Temporal Gating branches, followed by an MHA block and a Soft Mix addition operation to produce the final output.}
\end{figure}

\subsection{Impulse-Trend Synergistic Attention (ITSA)}
\label{subsec:itsa}

In dyadic conversations, human gestures physically decouple into two distinct properties: smooth, long-term kinematic trends and sparse, instantaneous rhythmic impulses driven by speech beats. However, relying purely on standard spatial-temporal Transformers incurs prohibitive computational overhead and causes severe semantic entanglement and over-smoothing.

To explicitly decouple ``what moves'' and ``when it moves'', we propose the Impulse-Trend Synergistic Attention (ITSA). As illustrated in \textbf{Fig.~\ref{fig:itsa}}, ITSA adds two lightweight, complementary 1D-convolutional gating streams to a global MHA branch. Its main contribution is representational rather than a reduction in block-level latency. Given the hidden feature $\mathbf{H} \in \mathbb{R}^{B \times D \times T}$, ITSA processes it as follows:

\paragraph{Trend Feature Gating (``What moves''):} This stream captures global anatomical saliency. We aggregate the temporal dynamics by applying both Average Pooling and Max Pooling along the time dimension $T$, and sum their results. This vector is processed by a pointwise convolutional bottleneck block $\mathcal{F}_{\text{feat}}$ (comprising $\text{Conv}_{1\times1} \rightarrow \text{BN} \rightarrow \text{ReLU} \rightarrow \text{Conv}_{1\times1}$) to generate a channel-wise mask $\mathbf{g}_{\text{feat}} \in \mathbb{R}^{B \times D \times 1}$:
\begin{equation}
    \mathbf{g}_{\text{feat}} = \sigma \left( \mathcal{F}_{\text{feat}} \left( \text{AvgPool}_T(\mathbf{H}) + \text{MaxPool}_T(\mathbf{H}) \right) \right).
\end{equation}

\paragraph{Impulse Temporal Gating (``When it moves''):} This stream localizes sparse, audio-driven motion beats. We pool the hidden features along the channel dimension $D$ using both Average and Max Pooling, concatenate them (denoted as $\parallel$), and apply a temporal convolutional block $\mathcal{F}_{\text{time}}$ (comprising $\text{Conv}_{1\times7} \rightarrow \text{BN} \rightarrow \text{ReLU} \rightarrow \text{Conv}_{1\times1}$) to produce the temporal mask $\mathbf{g}_{\text{time}} \in \mathbb{R}^{B \times 1 \times T}$:
\begin{equation}
    \mathbf{g}_{\text{time}} = \sigma \left( \mathcal{F}_{\text{time}} \left( [ \text{AvgPool}_D(\mathbf{H}) \parallel \text{MaxPool}_D(\mathbf{H}) ] \right) \right).
\end{equation}
By employing a large temporal receptive field (kernel size $1 \times 7$), this gating mechanism effectively identifies rapid, beat-aligned transients in the acoustic-driven trajectory, providing a sparse temporal prior for the subsequent fusion stage.

\paragraph{Learnable Soft-Mixture with MHA:} The two masks simultaneously modulate the original input feature via element-wise multiplication ($\mathbf{H} \odot \mathbf{g}_{\text{feat}} \odot \mathbf{g}_{\text{time}}$). Crucially, while this dual-gating provides powerful local spatiotemporal decoupling, it may lack sufficient global receptive fields for long-term trajectory planning. To address this, we compute a parallel representation $\mathbf{H}_{\text{MHA}}$ using a standard Multi-Head Attention layer. Instead of relying on a naive sequential connection or a rigid residual addition, we introduce an adaptive \textit{learnable soft-mixture} mechanism (depicted as $\oplus$ in \textbf{Fig.~\ref{fig:itsa}}) to seamlessly fuse the local synergistic features with the global contextual features:
\begin{equation}
    \tilde{\mathbf{H}} = (1 - \alpha) \cdot (\mathbf{H} \odot \mathbf{g}_{\text{feat}} \odot \mathbf{g}_{\text{time}}) + \alpha \cdot \mathbf{H}_{\text{MHA}}
\end{equation}
where $\alpha = \sigma(w_{\text{mix}})$ denotes the sigmoid activation over a learnable mixing logit $w_{\text{mix}}$ optimized end-to-end. This adaptive soft-blending balances high-frequency speech-driven transients with low-frequency global stability, reducing the tendency of the downstream generator to produce out-of-distribution jitter.

\begin{figure}[htbp]
  \centering
  \includegraphics[width=1\linewidth]{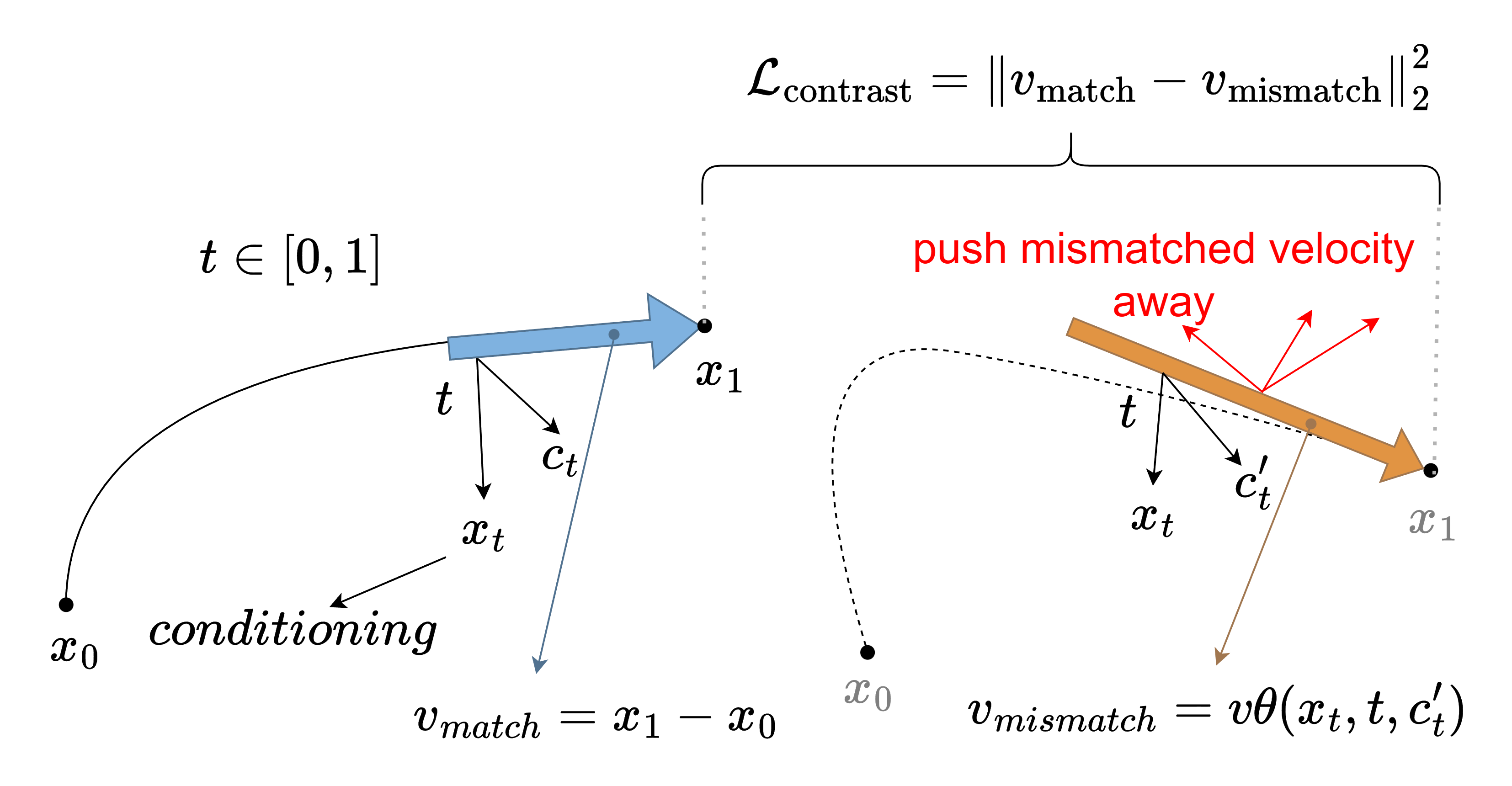}
  \caption{Conceptual audio--motion CFM at $d=0$: $x_1-x_0$ is the matched target direction, and the loss contrasts matched and shuffled-condition velocity predictions.}
  \label{fig:cfm} 
  \Description{A contrastive flow matching diagram comparing velocity fields conditioned on matched and mismatched audio-motion pairs. Red arrows indicate the repulsion applied to mismatched semantic conditions.}
\end{figure}

\subsection{Contrastive Flow Matching Objective}
\label{subsec:cfm}

To improve semantic alignment while supporting efficient inference, we optimize four complementary constraints.

\textbf{Standard Flow Matching.} Following the flow matching paradigm~\cite{Lipman2023FlowMatching}, let $x_0 \sim \mathcal{N}(0, I)$ and $x_1 \sim q(x)$ denote a clean ground-truth motion sequence. The optimal transport linear interpolant path is $x_t = (1-t)x_0 + t x_1$. Our flow estimator $v_\theta$ predicts the velocity field conditioned on the composite feature $c_t$; the base flow-matching examples use $d=0$. The standard loss is formulated as:
\begin{equation}
    \mathcal{L}_{\text{FM}} = \mathbb{E}_{t, x_0, x_1, c_t}\left[ 
      \left\| v_\theta(x_t, t, 0, c_t) - (x_1 - x_0) \right\|_2^2 
    \right].
\end{equation}

\textbf{Audio-Motion Contrastive Regularization.} Vanilla flow matching can produce insufficiently separated velocities when faced with ambiguous audio. To explicitly penalize semantically mismatched conditions, we introduce an intra-batch contrastive regularizer $\mathcal{L}_{\text{contrast}}$. Specifically, at the same latent state $x_t$, we compute the $L_2$ distance between the model's predicted velocity conditioned on the matched audio $c_t$ (acting as the prediction anchor, distinct from the ground-truth velocity) and the velocity conditioned on randomly shuffled audio $c'_t$. Maximizing this bounded distance encourages sensitivity to the correct conversational context:
\begin{equation}
    \mathcal{L}_{\text{contrast}} = \mathbb{E} \left[ \min\!\left(\left\| v_\theta(x_t, t, 0, c_t) - \text{sg}\left(v_\theta(x_t, t, 0, c'_t)\right) \right\|_2^2,\tau\right) \right],
\end{equation}
where $\text{sg}(\cdot)$ denotes stop-gradient and $\tau$ is the hard-clipping threshold. Detaching the mismatched prediction prevents unstable bidirectional divergence, while clipping bounds the contribution of each pair.

\textbf{Velocity Consistency Regularization.} To accelerate inference and improve trajectory stability with fewer sampling steps, we incorporate a consistency loss $\mathcal{L}_{\text{consistency}}$ following~\cite{Liu2025GestureLSM}. This objective encourages the velocity prediction to remain consistent across different integration step sizes. Specifically, the velocity estimated directly for a single large step of size $2\delta$ is trained to align with the average velocity of two consecutive smaller steps of size $\delta$. For brevity, let $v_s^{(d)}=v_\theta(x_s,s,d,c_t)$ for $s\in\{t,t+\delta\}$:
\begin{equation}
\begin{aligned}
    \mathcal{L}_{\text{consistency}}
    = \mathbb{E} \left[
      \left\| v_t^{(2\delta)}
      - \frac{v_t^{(\delta)} + v_{t+\delta}^{(\delta)}}{2}
      \right\|_2^2
    \right],
\end{aligned}
\end{equation}
where $x_{t+\delta} = x_t + \delta \cdot v_\theta(x_t,t,\delta,c_t)$ is the intermediate latent state. The two scalar inputs retain distinct meanings: $t$ specifies the current location on the flow path, while $d\in\{\delta,2\delta\}$ specifies the requested integration span.

\textbf{Full Objective.} Finally, we integrate the aforementioned losses with a Temporal Coherence penalty $\mathcal{L}_{\text{TC}}$ that minimizes the second-order temporal derivatives of the predicted velocity field to suppress high-frequency jitter. The comprehensive training objective is unified as:
\begin{equation}
    \mathcal{L}_{\text{total}} = \mathcal{L}_{\text{FM}} - g(s)\lambda_{\text{contrast}} \mathcal{L}_{\text{contrast}} + \mathcal{L}_{\text{consistency}} + \lambda_{\text{TC}} \mathcal{L}_{\text{TC}}.
\end{equation}
where $g(s)\in[0,1]$ is a warm-up gate indexed by training step $s$. The contrastive term is subtracted to maximize the bounded distance via standard gradient descent; after warm-up, $\lambda_{\text{contrast}}=0.05$, so flow matching remains the dominant objective.

\section{Experiment}
\subsection{Experimental Settings}

\textbf{Datasets.} We evaluate the proposed SynerGest framework on two extensive dyadic interaction datasets: the Audio2Photoreal dataset~\cite{Ng2024Audio2Photoreal} and Seamless Interaction~\cite{Agrawal2024Seamless}. Audio2Photoreal offers high-fidelity, studio-captured dyadic conversations paired with photorealistic Codec Avatar assets. The acoustic input comprises 48 kHz dual-channel audio, explicitly capturing the independent waveforms of both the speaker and the listener. Additionally, we utilize the Naturalistic subset of the Seamless Interaction dataset, which encompasses unscripted, spontaneous recordings. For Seamless, the release provides a single-channel track for each participant. Using the interaction-partner mapping, we pair the target and partner tracks, resample them to 48 kHz, align them to their common duration, and stack them as a dual-channel input.

\textbf{Implementation Details.} Our models are trained on a single NVIDIA RTX 4090 GPU. During inference, trajectory integration requires merely 8 ODE sampling steps.

\textbf{Evaluation Metrics.} We adopt four metrics to comprehensively evaluate our framework: Fr\'echet Gesture Distance (FGD $\downarrow$) for kinematic realism, Beat Consistency (BC $\uparrow$) for audio-motion synchronization, Diversity (Div $\uparrow$) for motion variance, and Inference Efficiency ($\mathcal{T}_{\text{inf}}$ $\downarrow$), defined as the average wall-clock inference time per sample.

\textbf{Baselines.} To rigorously demonstrate the superiority of our framework, we compare SynerGest against four recent state-of-the-art co-speech gesture generation methods. Specifically, we evaluate against three advanced \textit{monadic} frameworks—EMAGE~\cite{Liu2024EMAGE}, GestureLSM~\cite{Liu2025GestureLSM}, and TalkSHOW~\cite{Yi2023Holistic}—alongside one dedicated \textit{dyadic} baseline, the Audio2Photoreal (A2P) model~\cite{Ng2024Audio2Photoreal}. To ensure a fair comparison, all monadic baselines are adapted to our dyadic evaluation protocol under identical data processing conditions.

\subsection{Quantitative Results}
\label{subsec:quantitative}

\begin{table*}[t]
  \caption{Quantitative comparison on Audio2Photoreal and Seamless Interaction. $\mathcal{T}_{\text{inf}}$ is the average inference time per sample; best generated results are bold.}
  \label{tab:main_results}
  \renewcommand{\arraystretch}{0.86} 
  \footnotesize 
  \begin{tabular*}{\textwidth}{@{\extracolsep{\fill}} l l c c c c @{}}
    \toprule
    Dataset & Method & FGD $\downarrow$ & BC $\uparrow$ & Div $\uparrow$ & $\mathcal{T}_{\text{inf}}$ (s) $\downarrow$ \\
    \midrule
    \multirow{6}{*}{Audio2Photoreal} 
    & Ground Truth & -- & 0.703 & 12.15 & -- \\
    & EMAGE~\cite{Liu2024EMAGE} & 0.908 & 0.725 & 12.81 & 0.332 \\
    & GestureLSM~\cite{Liu2025GestureLSM} & 0.951 & 0.721 & \textbf{16.51} & \textbf{0.116} \\
    & TalkSHOW~\cite{Yi2023Holistic} & 3.824 & 0.674 & 13.97 & 0.410 \\
    & Audio2Photoreal (A2P)~\cite{Ng2024Audio2Photoreal} & 2.743 & 0.623 & 13.32 & 0.378 \\
    & \textbf{Ours} & \textbf{0.741} & \textbf{0.782} & 14.14 & 0.127  \\
    \midrule
    \multirow{6}{*}{Seamless Interaction} 
    & Ground Truth & -- & 0.722 & 14.13 & -- \\
    & EMAGE~\cite{Liu2024EMAGE} & 2.636 & 0.723 & 12.35 & 0.211 \\
    & GestureLSM~\cite{Liu2025GestureLSM} & 2.533 & 0.680 & \textbf{18.13} & \textbf{0.192} \\
    & TalkSHOW~\cite{Yi2023Holistic} & 2.925 & 0.692 & 12.15 & 0.436 \\
    & Audio2Photoreal (A2P)~\cite{Ng2024Audio2Photoreal} & 2.735 & 0.635 & 12.93 & 0.432 \\
    & \textbf{Ours} & \textbf{2.380} & \textbf{0.753} & 13.50 & 0.231 \\
    \bottomrule
  \end{tabular*}
\end{table*}

Table~\ref{tab:main_results} summarizes the quantitative comparison of our framework across two benchmarks.

\textbf{Performance on Audio2Photoreal (A2P).} On the A2P dataset featuring high-quality studio dyadic interactions, SynerGest demonstrates clear superiority. Specifically, our method reduces the FGD to 0.741, outperforming the strongest competitor EMAGE (0.908) by 18.4\% and realizing a 73.0\% error reduction over the A2P baseline. This improvement highlights the efficacy of our Flow Estimator in producing stable, semantically grounded trajectories. Furthermore, the ITSA module effectively captures rhythmic impulses by separating them from kinematic trends, yielding an outstanding BC score of 0.782.

\textbf{Analysis of Motion Diversity (Div).} Notably, while Ges\-ture\-LSM achieves the highest numerical Diversity (Div), we attribute this partly to less constrained diversity that includes kinematically noisy samples, as evidenced by its lower BC scores and qualitative jitter. In contrast, SynerGest explicitly pushes away mismatched semantic pairs via the CFM objective to avoid mode collapse, achieving Div scores (14.14 and 13.50) that closely mirror the natural variance of the Ground Truth. This ensures that the generated gestures are expressive and contextually appropriate without introducing erratic artifacts.

\textbf{Performance on Seamless Interaction.} SynerGest maintains leading performance on the more challenging Seamless dataset, achieving the lowest FGD of 2.380. This proves our framework's robustness in capturing accurate human motion in the wild without over-smoothing. Notably, our method sets a new state-of-the-art for beat alignment on this dataset (BC: 0.753), demonstrating that dual-channel audio injection and ITSA's impulse-trend decomposition remain highly effective even in unconstrained conversational environments. 

\textbf{Inference Efficiency ($\mathcal{T}_{\text{inf}}$).} SynerGest's system-level speed primarily comes from the streamlined 8-step ODE solver rather than from reduced attention-block cost. The reported ITSA module retains a global MHA branch and therefore has block-level latency comparable to standard attention. Even with this global branch, SynerGest achieves 0.127s on A2P and 0.231s on Seamless---3.0$\times$ and 1.9$\times$ faster than the A2P baseline, respectively. GestureLSM remains marginally faster because of its vanilla FM backbone.


\subsection{Qualitative Results}
\label{subsec:qualitative}

\begin{figure*}[t]
  \centering
  \includegraphics[width=\textwidth, height=8.5cm, keepaspectratio]{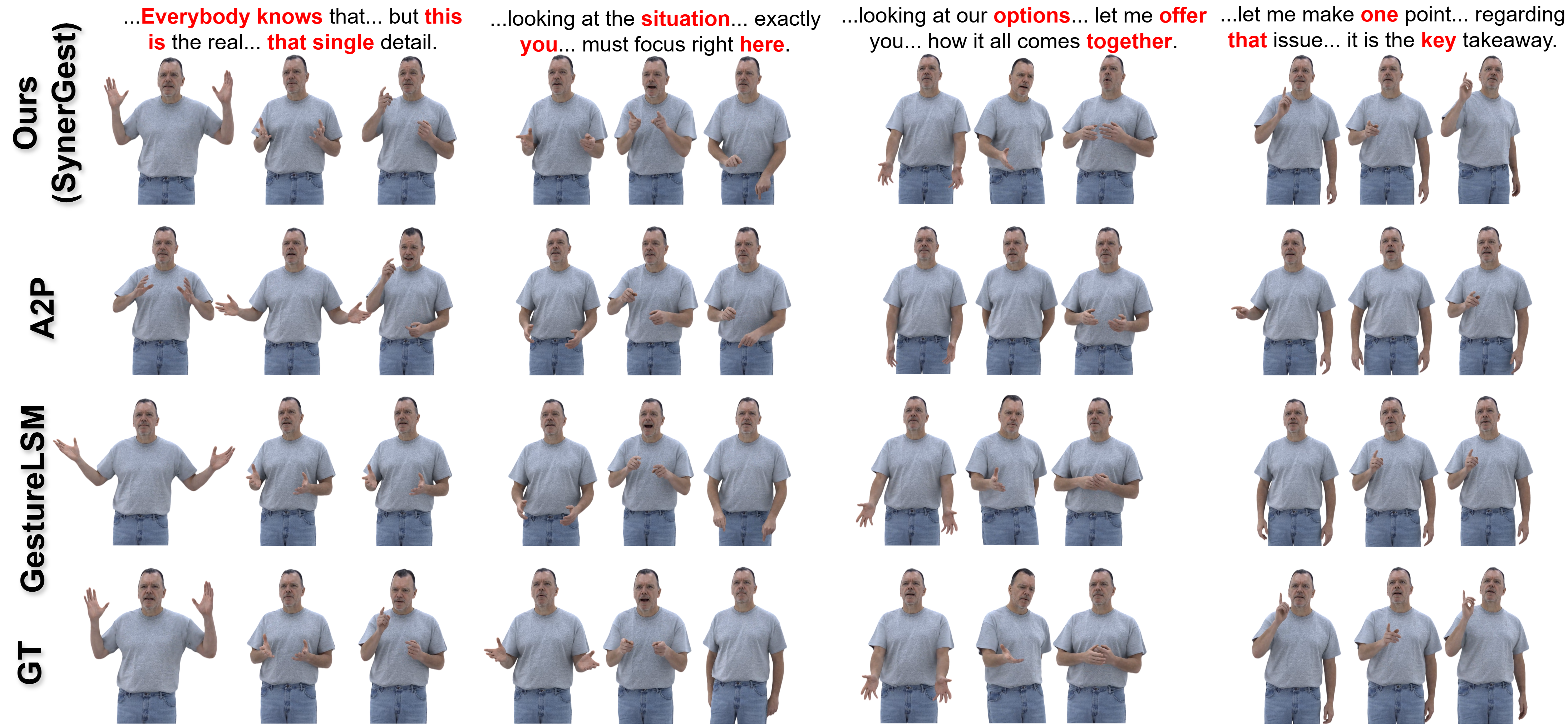} 
  \caption{Qualitative comparison with baselines and GT. Red highlights show that SynerGest captures semantically aligned gestures while avoiding mean-pose regression and kinematic jitter.}
  \label{dxfx}
  \Description{Frame sequences comparing ground truth, A2P, GestureLSM, and SynerGest. Red annotations highlight semantically aligned gestures and visible differences in pose quality and temporal stability.}
\end{figure*}

\textbf{Qualitative Comparisons.} We prioritize comparisons against A2P (the foundational dyadic baseline) and GestureLSM (FM-based), establishing a strictly comparable setting to demonstrate our improvements over vanilla FM. As shown in \textbf{Fig.~\ref{dxfx}}, SynerGest effectively mitigates the \textit{mean-pose regression} prevalent in A2P. While A2P often retreats into passive, semantically incorrect "hands-behind-the-back" postures, SynerGest accurately replicates the ground-truth semantic trajectory, such as evolving from an open-handed introduction to a definitive hand-clasp.

Furthermore, while Ges\-ture\-LSM ex\-hib\-its no\-tice\-able high-fre\-quen\-cy jit\-ter, SynerGest en\-sures su\-pe\-ri\-or flu\-id\-i\-ty through continuous trajectory integration and $\mathcal{L}_{\text{TC}}$. Crucially, our framework captures sparse, rapid semantic impulses that baselines often overlook. In sequences requiring decisive index-finger gestures synchronized with acoustic onsets (\textbf{Fig.~\ref{dxfx}}), SynerGest executes crisp transients where Ges\-ture\-LSM maintains ambiguous arm postures. These observations validate the ITSA module's ability to disentangle sharp speech-driven beats from continuous kinematic trends without introducing jitter.

\begin{figure}[t]
  \centering
  \includegraphics[width=1\linewidth, keepaspectratio]{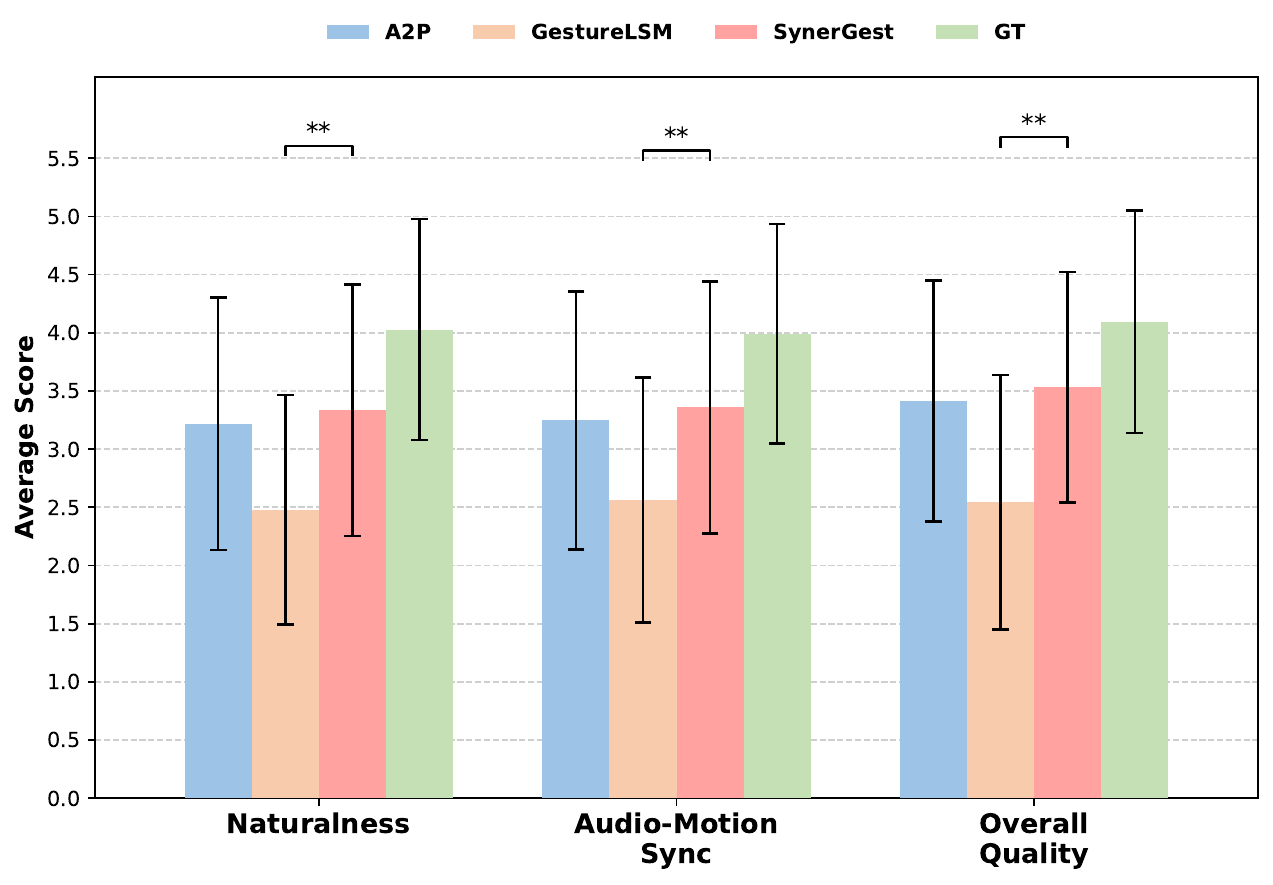} 
  \caption{User-study MOS with 95\% CIs. SynerGest significantly outperforms both baselines.}
  \label{fig:user_study}
  \Description{A chart of mean opinion scores with 95 percent confidence intervals for naturalness, audio-motion synchronization, and overall quality across SynerGest and baseline methods.}
\end{figure}

\textbf{User Study.} A formal user study with 45 participants using a 5-point Mean Opinion Score (MOS) scale confirms SynerGest's superiority (\textbf{Fig.~\ref{fig:user_study}}). Our framework leads in \textit{Naturalness} (3.33), \textit{Audio-Motion Sync} (3.36), and \textit{Overall Quality} (3.53). Wilcoxon signed-rank tests show significant improvements over GestureLSM ($p < 0.0001$ for all three dimensions) and A2P (Naturalness: $p=0.041$; Audio-Motion Sync: $p=0.024$; Overall Quality: $p=0.035$). This subjective performance further validates the overall efficacy of \textbf{SynerGest} in generating contextually responsive dyadic gestures.

\subsection{Ablation Studies}
\label{subsec:ablation}

To verify the effectiveness of our proposed modules, we perform experiments on the A2P and Seamless Interaction benchmarks. Following a subtractive ablation protocol, we begin with the full configuration and progressively disable specific components to analyze their necessity. Table~\ref{tab:ablation} shows the performance changes across four key dimensions.

\begin{table}[t]
  \caption{Component ablations on Audio2Photoreal and Seamless Interaction.}
  \label{tab:ablation}
  \renewcommand{\arraystretch}{0.86}
  \resizebox{\linewidth}{!}{
  \begin{tabular}{llccc}
    \toprule
    Dataset & Model Variant & FGD $\downarrow$ & BC $\uparrow$ & Div $\uparrow$ \\
    \midrule
    \multirow{4}{*}{Audio2Photoreal} 
    & \textbf{Full Model (Ours)} & \textbf{0.741}& \textbf{0.782}& 14.14\\
    \cmidrule{2-5}
    & w/ Diffusion (replace FM) & 2.170& 0.639& 13.38\\
    & w/o Contrastive Loss (Vanilla FM) & 1.440& 0.693& 13.80\\
    & w/o ITSA (use standard MHA) & 1.334& 0.712& \textbf{17.78}\\
    \midrule
    \multirow{4}{*}{Seamless Interaction} 
    & \textbf{Full Model (Ours)} & \textbf{2.380}& \textbf{0.753}& 13.50\\
    \cmidrule{2-5}
    & w/ Diffusion (replace FM) & 3.219& 0.726& 7.026\\
    & w/o Contrastive Loss (Vanilla FM) & 3.268& 0.716& 9.841\\
    & w/o ITSA (use standard MHA) & 3.502& 0.633& \textbf{14.59}\\
    \bottomrule
  \end{tabular}
  }
\end{table}

\textbf{Flow Matching vs. Diffusion.} 
We replace our Flow Matching backbone with a Denoising Diffusion Probabilistic Model (DDPM) utilizing the DDIM~\cite{Song2021DDIM} sampling strategy. To provide a rigorous and competitive baseline, the diffusion variant is evaluated with $100$ sampling steps (\textit{DDIM-100}), whereas SynerGest requires only $8$ ODE steps. As shown in Table~\ref{tab:ablation}, despite the $12.5\times$ higher sampling budget allocated to the diffusion baseline, it still suffers a significant degradation in FGD, worsening from 0.741 to 2.170 on A2P, and exhibits poor beat synchronization (BC: 0.639). On the Seamless dataset, its Diversity score drops to 7.026, indicating reduced coverage of the motion distribution under this experimental setting.

\textbf{Contrastive Flow Matching (CFM) vs. Vanilla FM.} 
Removing the contrastive objective turns the model into a Vanilla FM. As a result, the FGD worsens significantly from 0.741 to 1.440 on A2P. Without the explicit separation between matched and mismatched audio conditions, conditional discrimination becomes weaker. On the Seamless dataset, Diversity drops from 13.50 to 9.841, consistent with a stronger tendency to generate safe, over-smoothed average gestures instead of clear, interactive motions.

\textbf{ITSA vs. Standard Attention.} We replace the entire ITSA module (including both the dual-gating streams and the learnable soft-mixture mechanism) with a standard Multi-Head Attention layer. While this variant obtains abnormally high Diversity scores (17.78 on A2P and 14.59 on Seamless), its motion quality heavily deteriorates, with FGD worsening to 1.334 (A2P) and 3.502 (Seamless). Consistent with our analysis in Sec.~\ref{subsec:quantitative}, standard attention fails to separate sharp speech impulses from smooth motion trends, causing the network to translate acoustic noise into unnatural structural jitter, which artificially inflates the Diversity metric without adding semantically meaningful variation. ITSA explicitly decouples rhythm and trends to preserve physical responsiveness and stability.

\section{Conclusion}
\label{sec:conclusion}

In this paper, we present \textbf{SynerGest}, a novel generative framework advancing holistic co-speech gesture synthesis from isolated monadic generation to dyadic interactive embodiment. To improve the conditional sensitivity of vanilla Flow Matching under ambiguous conversational audio, we adapt Contrastive Flow Matching (CFM) to dual-audio conditioning. By explicitly maximizing the $L_2$ distance between velocity predictions under matched and shuffled conversational contexts, CFM improves conditional separation and reduces the tendency to generate generic, over-smoothed gestures. Furthermore, we propose the Impulse-Trend Synergistic Attention (ITSA) module to model continuous kinematic trends and sparse, speech-driven rhythmic impulses via complementary dual-gating streams and an adaptive soft-mixture mechanism. Extensive evaluations on Audio2Photoreal and Seamless Interaction demonstrate that SynerGest achieves state-of-the-art performance in both kinematic realism and rhythmic synchronization under our protocol. By reconciling rhythmic impulses with kinematic stability, SynerGest provides a viable framework for modeling the reactive behavioral dynamics of human conversation. 

\begin{acks}
This work was supported by the Key Research and Development Program of
Yunnan Province under Grant Nos. 202502AD080017 and 202402AA310056.
\end{acks}

\bibliographystyle{ACM-Reference-Format}
\bibliography{sample-base}

\end{document}